\documentclass[journal]{IEEEtran}

\ifCLASSOPTIONcompsoc

  \usepackage[nocompress]{cite}
\else
  \usepackage{cite}
\fi
\usepackage{multirow}
\usepackage{amsmath}
\usepackage{mathtools}
\usepackage{amssymb}
\usepackage{breqn}
\usepackage{kantlipsum}
\usepackage{algorithm}
\usepackage{caption}
\usepackage{subcaption} %%% 
\usepackage{graphicx}
\usepackage[algo2e]{algorithm2e}
\usepackage{array}
\newcolumntype{x}[1]{>{\centering\arraybackslash\hspace{0pt}}p{#1}}
\newcolumntype{y}[1]{>{\lecentering\arraybackslash\hspace{0pt}}p{#1}}
\usepackage{graphicx} % Allows including images
\RequirePackageWithOptions{multicol}
\usepackage{stfloats}

\ifCLASSINFOpdf
 
\else

\fi

\begin{document}

\title{Image Clustering using an Augmented Generative Adversarial Network and Information Maximization}
\author{Foivos~Ntelemis,
        Yaochu~Jin, \emph{Fellow}, \emph{IEEE},
        Spencer~A.~Thomas
\thanks{This project is funded by an EPSRC industrial CASE award (number 17000013) and Department for Business, Energy and Industrial Strategy through the National Measurement System (122416). (\textit{Corresponding author: Yaochu Jin})}
\thanks{F. Ntelemis and Y. Jin are with the Department of Computer Science, University of Surrey, Guildford, GU2 7XH, United Kingdom. (Email: \{f.ntelemis; yaochu.jin\}@surrey.ac.uk)}
\thanks{S. A. Thomas is with the National Physical Laboratory, Teddington, TW11 0LW, United Kingdom. (Email: spencer.thomas@npl.co.uk)}
%and Technology, Shanghai, 200237, P. R. China (e-mail: \{donghan;duwei0203;wldu\}@ecust.edu.cn).}% <-this % stops a space
\thanks{Manuscript received xx, 2020; revised xx, 2020.}
}

\maketitle

\IEEEtitleabstractindextext{%

\begin{abstract}

Image clustering has recently attracted significant attention due to the increased availability of unlabelled datasets. The efficiency of traditional clustering algorithms heavily depends on the distance functions used and the dimensionality of the features. Therefore, performance degradation is often observed when tackling either unprocessed images or high-dimensional features extracted from processed images. To deal with these challenges, we propose a deep clustering framework consisting of a modified generative adversarial network (GAN) and an auxiliary classifier. The modification employs Sobel operations prior to the discriminator of the GAN to enhance the separability of the learned features. The discriminator is then leveraged to generate representations as the input to an auxiliary classifier. An adaptive objective function is utilised to train the auxiliary classifier for clustering the representations, aiming to increase the robustness by minimizing the divergence of multiple representations generated by the discriminator. The auxiliary classifier is implemented with a group of multiple cluster-heads, where a tolerance hyper-parameter is used to tackle imbalanced data. Our results indicate that the proposed method significantly outperforms state-of-the-art clustering methods on CIFAR-10 and CIFAR-100, and is competitive on the STL10 and MNIST datasets. 

\end{abstract}


}

\IEEEdisplaynontitleabstractindextext

\IEEEpeerreviewmaketitle

\section{Introduction}\label{sec:introduction}

\IEEEPARstart{N}{owadays} an increasing amount of large-scale unlabelled visual data has been made available. However, most existing learning algorithms are either designed for supervised learning where the ground truth is known, or for unsupervised learning using distance measurements. As for supervised approaches, manual annotation is a time-consuming task and can provide a temporary solution only since the labels correspond exclusively to a specific dataset \cite{Lin2014MicrosoftCC}. \par

Clustering tasks have been widely studied from feature extraction \cite{Boutsidis2009UnsupervisedFS,DBLP:books/crc/aggarwal13/AlelyaniTL13}, to grouping algorithms \cite{MacQueen1967SomeMF,DBLP:conf/iccv/ComaniciuM99,1017616} and distance measurements \cite{Xing2002DistanceML,XIANG20083600,Loohach_effectof,Singh_k-meanswith}. Regarding the grouping algorithms, a range of breakthrough approaches from different perspectives has been proposed in the literature, such as centroid \cite{MacQueen1967SomeMF,BEZDEK1984191,DBLP:conf/iccv/ComaniciuM99}, hierarchical \cite{Heller2005BayesianHC,Williams1999AMA} and graph-based \cite{10.1007/978-3-642-33718-5_31}. However, their effectiveness heavily depends on the dimensionality of the training data as well as the defined distance functions \cite{Loohach_effectof}. Therefore, the usefulness of most existing clustering approaches is limited when handling high-dimensional data \cite{Steinbach2004, DBLP:conf/iccv/ChangWMXP17,5454426}. Several studies address these concerns by means of linear data reduction \cite{1015408, 1421871,Torre2006DiscriminativeCA}, which theoretically resolves the issue of the features scale. Decomposition methods such as principal component analysis (PCA) \cite{Hotelling_1933} and non-negative matrix factorization (NMF)\cite{Lee2000AlgorithmsFN} are restrictive when handling complex data types such as images, since the semantic information can be located in any position of the image and often a priori texture analysis is required \cite{AbdelHakim2006CSIFTAS,Masoudi2016ClassificationOC}.
One may face similar challenges when using distance functions, since it is difficult to define such a measure \cite{DBLP:conf/iccv/ChangWMXP17,5454426}. \par

Recent studies address dimensionality reduction and semantic issues with the help of deep learning techniques that replace the traditional data reduction and feature extraction methods \cite{Yang2016JointUL,DBLP:journals/corr/abs-1807-05520,10.1007/978-3-319-70096-0_39, 10.1007/978-3-642-21735-7_7}. These studies demonstrate that neural network models are powerful tools for learning representations and extracting features. Despite their success and significant advantages, the clustering process is frequently applied using traditional algorithms, which often leads to degenerate solutions or leaves no space for further improvement \cite{DBLP:journals/corr/abs-1807-05520,pmlr-v48-xieb16}. Few studies propose comprehensive solutions \cite{Yang2016JointUL,DBLP:conf/iccv/ChangWMXP17,DBLP:conf/dagm/HausserPGAC18,DBLP:journals/corr/abs-1807-06653} based on a deep convolutional architecture, which, however, often are either applicable to well-known benchmark datasets only, or computationally very intensive. Recently, generative adversarial networks (GANs) have attracted significant attention \cite{NIPS2014_5423,radford2015unsupervised,donahue2017adversarial} in representation learning in an unsupervised manner, either for generating synthetic samples or for distinguishing between the training set and fake samples.\par

In order to deal with these challenges, this work introduces a deep learning method composed of two separate training phases. At the first phase, a modified generative adversarial network (GAN)~\cite{NIPS2014_5423} is trained to learn the representation of the dataset in a self-supervised learning manner. The Sobel filters are introduced prior to the discriminator of the GAN proposed in \cite{radford2015unsupervised}, which aims to enhance the capacity of the discriminator in learning image representation and extracting meaningful features. Afterwards, the generator part is discarded and the discriminator implements a feature extraction pipeline as an input into an auxiliary classifier. At the second phase, the auxiliary classifier consists of an additional neural net, which is trained separately from the GAN according to a modified clustering objective originally introduced for information maximizing self-augmented training (IMSAT) \cite{Hu2017}. We extend the loss function in the original IMSAT, which utilizes a single cluster-head, by adopting a set of multiple cluster-heads, each being trained independently. We therefore parameterize the initial function to deal with the combined cluster-heads, allowing a higher degree of flexibility leading to an improvement in clustering accuracy as demonstrated by the experimental results. Furthermore, we introduce a penalty term to minimize the divergence across invariant generated representations derived by the GAN framework.

\par 
The main contributions of this work are summarized below.
\begin{itemize}

\item A comprehensive deep learning framework is proposed for solving binary pairwise clustering problems for image data without relying on domain-specific augmentation operations.

\item The Sobel filter is introduced into the discriminator of the GAN framework to enhance the ability of the discriminator for learning disentangled representations.

%The Sobel filters are introduced into the GAN framework to improve the representation learning ability of the discriminator of the GAN.

\item An extended auxiliary classifier is proposed, which is optimised via a modified loss function by adding a penalty term measuring the divergence across invariant representation derived by the discriminator. 

\end{itemize}   
% Our results indicate that the proposed method significantly outperforms state-of-the-art clustering methods on CIFAR-10 and CIFAR-100, and is competitive on the STL and MNIST datasets. 
% my model doesnt have the highest score in STL, will this need to be included above?
\par The rest of the paper is organized in four main sections. Related work including a brief literature review is presented in Section II. The proposed end-to-end deep clustering method is detailed in Section III. Section IV describes the experimental results and comparisons across a range of competitive methods. Conclusion and future work are given in Section V.

\section{Related work}

Data clustering has attracted much research effort over the last decades \cite{Yang2016JointUL , DBLP:conf/iccv/ChangWMXP17 , DBLP:journals/corr/jiangZTTZ16, Hu2017, DBLP:conf/dagm/HausserPGAC18, DBLP:journals/corr/abs-1807-06653, pmlr-v48-xieb16,Reynolds2009GaussianMM,DBLP:conf/iccv/ComaniciuM99,1017616}. In general, traditional clustering methods are divided into two main categories: 1) Generative methods, such as Gaussian mixture models \cite{Reynolds2009GaussianMM}, learn the class-conditional probability of each class individually, and the prediction process is formulated with the direct application of Bayes' Theorem, where the prior and posterior probabilities are derived \cite{murphy2013machine}. On the other hand, the discriminative models explicitly learn posterior probabilities by defining hyperplanes and decision boundaries in the training samples \cite{murphy2013machine}.
\par

Over the last decade, a large body of research on deep learning based image clustering methods has been reported \cite{Yang2016JointUL , DBLP:conf/iccv/ChangWMXP17 , DBLP:journals/corr/jiangZTTZ16, Hu2017, DBLP:conf/dagm/HausserPGAC18, DBLP:journals/corr/abs-1807-06653, pmlr-v48-xieb16}. Early studies proposed the combination of a deep autoencoder with k-means in order to eliminate the linearity \cite{DBLP:journals/corr/YangFSH16,pmlr-v48-xieb16}. Xie et al. \cite{pmlr-v48-xieb16} presented a comprehensive solution named deep embedding clustering (DEC), where an autoencoder projects the training samples into a lower domain space via a bottleneck layer. Regardless of the functionality of the neural networks, the clusters' centroids are initialized with the help of k-means and a loss function depending on the distance measurement. However, DEC performance is limited to image data, since it requires the utilization of histogram oriented gradients (HOG) \cite{Dalal2005HistogramsOO} for relevant feature extraction. Inspired by the agglomerative methods, joint unsupervised learning (JULE)\cite{Yang2016JointUL} was proposed to tackle image clustering from a different perspective, since distance functions are difficult to define. Despite its success, the applicability of JULE is restricted due to the recurrent framework which requires high computational cost and memory sources \cite{DBLP:conf/dagm/HausserPGAC18}. In deep subspace clustering networks (DSC-Nets), an autoencoder architecture is leveraged to express a lower domain subspace. The clustering phase is achieved through a self-expressing matrix, which is of a dimension of $\mathbb{R}^{n \times n}$, where $n$ denotes the size of training set. Although the idea is very interesting, its applicability to large datasets is limited due to memory restrictions.  \par
More recently, mutual information theory was introduced into deep learning based clustering methods \cite{DBLP:journals/corr/abs-1807-06653,Hu2017,hjelm2019learning}. Gomes et al \cite{Gomes2010} initially presented regularized information maximization (RIM), a discriminative approach where a conditional entropy term is used to maximize the confidence in clustering assignment of each element to the relevant class. The second term of marginal entropy naturally regularizes the model to evenly assign the training samples in relevant clusters. In addition to the entropy terms, an extra regularization term is proposed \cite{Gomes2010} to enhance the parametric model's ability to identify sensible decision boundaries within the given set. \par
An application of RIM to a deep method was proposed by Hu et al. \cite{Hu2017}, which introduced IMSAT, with the direct application of two entropy terms as a loss function. In IMSAT, the additional penalty term is replaced by an alternative self-regularization method proposed for virtual adversarial training (VAT) by Miyato et al \cite{Miyato2017VirtualAT}. The combination of RIM and VAT in a multilayer perceptron net produces impressive clustering capability on the MNIST dataset \cite{Hu2017}. Note that in IMSAT, a prior feature extraction is required by an external pre-trained model when dealing with multi-dimensional images (RGB) or an affine transformation together with VAT. Affine transformation is a stochastic process, therefore its performance heavily depends on the augmented function and their hyper-parameters, such as the translation range, the degree of rotations, and the scales of color jittering. \par

Studies that exclusively apply image transformation/augmentation techniques to clustering have also been reported in \cite{DBLP:journals/corr/abs-1807-06653,DBLP:journals/corr/abs-1807-05520,DBLP:journals/corr/DosovitskiySRB14}. In invariant information clustering (IIC) \cite{DBLP:journals/corr/abs-1807-06653}, mutual information is explicitly maximized with respect to the original data and several augmented transformed versions. A set of cluster heads and a set of over-cluster heads are developed to improve the robustness in assigning the corresponding cluster index. Deep InfoMax (DIM)~\cite{hjelm2019learning} is essentially an implicit measure of mutual information, maximizing the lower bound between a group of spatial representations and a global representation. Despite the encouraging results, DIM requires to compute multiple complex estimations. \par

In contrast to \cite{hjelm2019learning}, the measurement of mutual information's lower bound, particular in Donsker-Varadhan \cite{c52c83e0b02c4746a5ea29b5cd44fd00}, may have statistical limitations, as pointed out by McAllester and Statos \cite{mcallester2019formal}. Here, we employ a GAN framework, originally designed for learning the data distribution for feature extraction. Furthermore, during the clustering process, we extend the functionality of IMSAT to cluster the learned features without the implementation of an external model. Inspired by \cite{DBLP:journals/corr/abs-1807-06653}, we adopt a large number of probabilistic outputs and over-cluster heads, which is combined with a tolerance hyper-parameter in the proposed methodology to allow for flexibility on imbalanced datasets and to achieve better clustering accuracy. Details of the implementation are provided in Section \ref{sec:method}.

\section{Method}
\label{sec:method}

\begin{figure}[t!]
    \centering
    \includegraphics[width=0.4\textwidth]{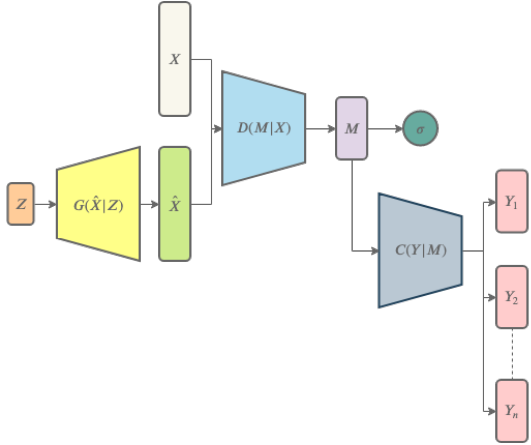}
    \caption{Diagram of the proposal framework for clustering multi-dimensional datasets, where $G$ and $D$ stands for generator and discriminator model, respectively. $M$ denotes the discriminator's output which is directed to auxiliary classifier. $C$ represents the auxiliary classifier and $Y_{i}$ the multi-cluster heads. }
    \label{fig:Framework}
\end{figure}

In this section, we outline the general framework of the proposed deep framework for image clustering, as illustrated in Fig. \ref{fig:Framework}. 
Initially, we make an assumption that for each given element in the unlabeled dataset $x \in X$, a binary pairwise relation holds to a discrete finite set such as $X \rightarrow Y $. In this case, $y \in \{1, 2,...,k\}$ and  $k$ is a predefined hyper-parameter which indicates the number of classes in the set of $X$. 
\par
The goal of our approach is to define a parametric model that satisfies the described relation $f_{\theta}: X \rightarrow Y$. In this work we assume that $X$ is a dataset comprised of multi-dimensional images. Our clustering process consists of two learning stages: 
\begin{itemize}
    \item \textbf{Learning deep representation:} A convolutional GAN \cite{NIPS2014_5423} is implemented to learn the distribution of the data to be clustered. This way, bottleneck feature extraction is achieved by leveraging the last layer of the discriminator model.
    \item \textbf{An Auxiliary Classifier:} An auxiliary classifier network is introduced at the output of the discriminator. The parameters of this additional network are optimised by applying IMSAT \cite{Hu2017}, a combination of the loss function of RIM \cite{Gomes2010}, with an extra penalty term for virtual adversarial training \cite{Miyato2017VirtualAT}.
\end{itemize}
In the following, we detail these training phases. 

\subsection{Learning Deep Representation}
The GAN framework \cite{NIPS2014_5423} was proposed as a straightforward technique for training deep generative models. The architecture consists of two main components: 1) a generator denoted as $G(z;\theta_g$), where $G$ is a multi-layer model, differentiable in all points with its input being a prior noise $z$ and an output conditioned on parameters $\theta_g$; and \ 2) a discriminator defined as $D(x;\theta_d)$, where $x$ denotes an input tensor with respect to the model's parameters $\theta_d$. Both models are optimised simultaneously with back-propagation via a minmax game, with the intention of approximating the parametric density $p_g(x)$ to the real distribution $p_{r}(x)$. The discriminator aims to distinguish the source of the input element by assigning the relevant probability. More formally, the minmax training function $V(D,G)$ of an adversarial model can be expressed as follows \cite{NIPS2014_5423}:

\begin{equation}
\begin{split}
\min_{G}\max_{D}V(D,G)&=E_{x \backsim p_{\text{data}}(x)}[\text{log}(D(x))] \\
&+ E_{z \backsim p_{z}(z)}[\text{log}(1-D(G(z))]
\end{split}
\end{equation}

\begin{figure*}[t]
    \centering
    \includegraphics[width=1\textwidth]{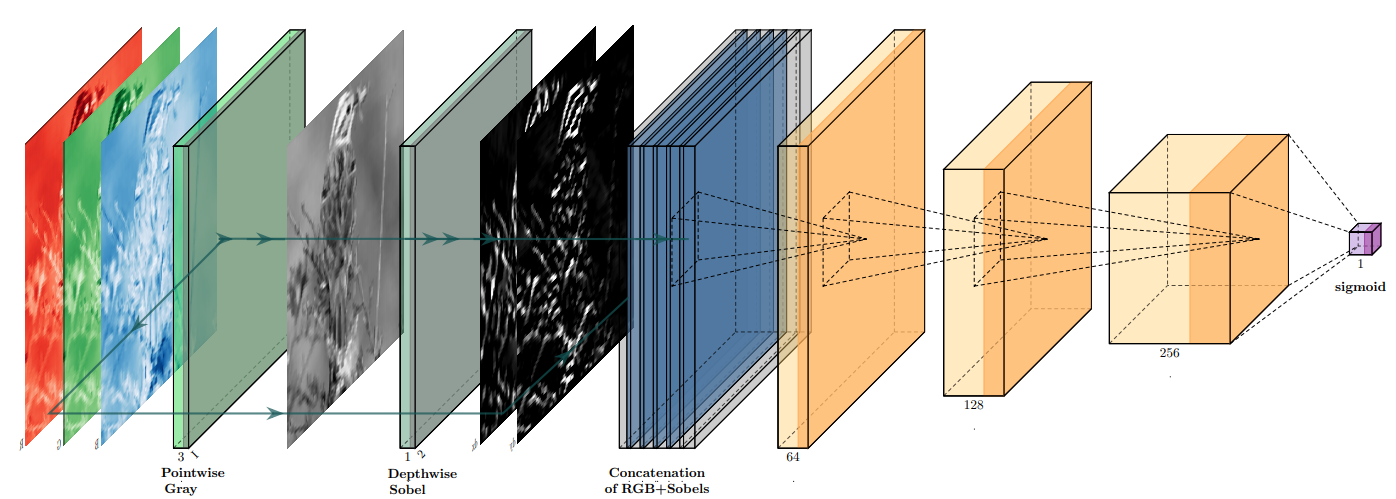}
    \caption{An illustration of the modified discriminator of the GAN framework in which two convolutional layers operate in front of the discriminator input. Initially, the three image layers (RGB) are converted into a gray-scale version via the application of pointwise layer. Afterwards a depthwise convolutional with a constant kernel implements the two directions of the Sobel's operations. The generated domain is concatenated with the original input and forwarded to the discriminator model.}
    \label{fig:Discriminator Structure}
\end{figure*}

Several studies have introduced alternative methods for training GANs \cite{DBLP:journals/corr/GulrajaniAADC17, arjovsky2017wasserstein, mao2016squares, DBLP:journals/corr/ZhaoML16}. Differing from the methodologies presented in these studies which mainly consider the quality of the generated sample, this work focuses on learning the distribution of training samples via a self-training process. The discriminator model is transformed into a feature extraction pipeline for clustering. Hence, the training strategy is not chosen with respect to the quality of generated samples; instead, it concentrates on the performance of the discriminator. For this reason, we adopt the deep convolutional generative adversarial network (DCGAN) \cite{radford2015unsupervised}, whose discriminator has been demonstrated to have a strong capability of extracting relevant features in imaging data \cite{radford2015unsupervised,Krizhevsky09learningmultiple}. Additionally, in \cite{article}, the discriminator is utilised by extracting features from each convolution layer, and a similar approach is adopted in \cite{8328474} for hyper-spectral image dataset. \par
Compared to the original structure of DCGAN, the discriminator of the GAN proposed in this work is forced to capture information such as saturation and colours by applying convolutional operations to the original image). Here, we apply the Sobel operators prior to the discriminator, which encourages the discriminator to capture as much detail as possible in terms of edges and shapes \cite{DBLP:journals/corr/abs-1807-05520,DBLP:journals/corr/abs-1807-06653,bojanowski2017unsupervised}. The Sobel operator begins by applying a simple pointwise convolutional layer which converts the red green blue (RGB) input into gray-scale. This is followed by a depthwise convolutional layer with predefined Sobel constant weights in both directions ($dx$,$dy$). The produced edges are concatenated with the original RGB sample and inputted to the discriminator model, as shown in Fig. \ref{fig:Discriminator Structure}. Since in both parts all layers are fully differentiable, the generator model can also be optimised in a similar way to back-propagation. This is also found to boost the already competitive learning performance of GANs. \par
The new domain space of the extracted features is flattened and defined as $M$  hereafter. 
To avoid extremely large values in the generated domain $M$, a $L_1$ regularization is added to the loss function in Equation (1) when training the flattened layer prior to the activation function of the discriminator. The regularization term is defined as $L_1(M)= \sum \mathrm{max}(|M|-20,0)$ to penalize the output of the flatten layer if its value is outside [-20, 20].

\subsection{Auxiliary Classifier}

Further to the above architecture, we train an auxiliary classifier network to cluster the features extracted by the trained discriminator. Our clustering strategy is based on IMSAT \cite{Hu2017} which implements RIM \cite{Gomes2010}.

Before explaining the training process, we briefly describe the mutual information and its application in RIM. Mutual information is defined by

\begin{dmath}
I(X;Y) = H(Y) - H(Y|X),
\label{eq:MI}
\end{dmath}
which encourages the increased discrepancy between the two entropy terms for different perspectives. According to the conditional entropy $H(Y|X)$, the model is driven to reduce the uncertainty by assigning the corresponding class index with higher confidence. This is achieved by minimizing the conditional entropy. On the other hand, the aim of maximizing the marginal entropy $H(Y)$ is to evenly distribute the model's class assignment. This avoids degenerate solutions, which are often observed in clustering tasks. According to information theory \cite{Cover:1991:EIT:129837}, the marginal entropy follows an upper bound $H(Y)\leq \text{log}(k)$. Therefore, instead of maximizing the marginal entropy, this can be transformed into minimization, where the following expressions are applicable \cite{Cover:1991:EIT:129837}:

\begin{dmath}
D_{KL}(p(y)||u) = \sum p(y)\text{log} \frac{p(y)}{u(y)} = \text{log}(k) - H(Y)
\label{eq:equality}
\end{dmath}

where $D_{KL}$ denotes to Kullback–Leibler divergence function, and $u$ is a uniformly distributed prior. Here, we combine the conditional entropy and Equation (\ref{eq:equality}). Consequently, we aim to minimize the following term as the RIM part of the objective function \cite{Hu2017}.

\begin{dmath}
I_{\psi}(M,Y)=\mathrm{max}([D_{KL}(p_{\psi}(y)||u(y))  - \delta],0) + H_{\psi}(Y|M)
\label{eq:minimize mi}
\end{dmath}
where $\delta \geq 0$ denotes an introduced tolerance factor which encourages the model's cluster distribution $p_{\psi}(y)$ to approximate the uniform prior within a defined tolerance level, and $\psi \in R$ is the auxiliary classifier parameters. The application of $\delta$-tolerance increases the model's capability of avoiding degenerate solutions, without which the model will be forced to balance the training samples, resulting in the assignment of trivial solutions in the case of imbalanced data. We found that by introducing a tolerance factor in optimizing the model's parameters using the gradient descent, we are able to enhance the accuracy of the model, as indicated by our experimental results presented in Subsection \ref{sub:auxiliary classifier}.

In the following, we elaborate the loss function to be used for training the auxiliary classifier. Here, we borrow and extend the idea of self-augmented training (SAT) initially introduced in IMSAT \cite{Hu2017}. The regularized term in IMSAT is described as \cite{Hu2017}:

\begin{dmath}
R_{SAT}(\psi;T) = \frac{1}{N}\sum_{n=1}^N{R_{SAT}(\psi;m_n,T(m_n))}
\label{eq:rsat}
\end{dmath}

where $T(m) = m + r$ denotes a transformed function with input $m$ and a perturbation $r$. 
The regularized function is defined using Kullback–Leibler  (KL) divergence, and theoretically the model's predictions will be invariant to data transformation.
In this work, we make use of the virtual adversarial training (VAT) \cite{Miyato2017VirtualAT} methodology as adopted in IMSAT as an implementation of $R_{SAT}$ :

\begin{equation}
    L_{adv}(x,\psi)=D_{KL}(p(y|m,\psi),p(y|m+r_{adv},\psi) 
\label{eq:radv}   
\end{equation}
\begin{equation}
\text{where } r_{adv} := \underset{r;\|r\|^2 \leq \epsilon}{\text{argmax}}\{D_{KL}(p_{\psi}(y|m),p_{\psi}(y|m+r)\}
%\end{split}
\label{eq:epsilon}
\end{equation}
where $r$ denotes random perturbations and $r_{adv}$ the adversarial perturbations. Similar to \cite{Miyato2017VirtualAT}, we approximate $r_{adv}$ based on the $L_2$ norm, i.e., $r_{adv} \approx \epsilon \frac{g}{\|g\|^2}$, where $g$ denotes the computed gradient with respect to the relevant point.
We adapt $\epsilon$ for both rounds of the VAT process to be proportional to the norm of each generated representation as $\epsilon_i = \alpha \cdot \|m_i\|^2$, where $\alpha$ is a constant scalar and $i$ is the index of the corresponding feature. We denote the constant scalar as $\alpha_{r}$ in the first round of random perturbation, and $\alpha_{adv}$ for the second round of adversarial perturbation.

The process was adapted from the clustering strategy in IMSAT implementing regularized information maximization \cite{Gomes2010} and VAT \cite{Miyato2017VirtualAT}. In this work, we aim to further enhance the robustness of the proposed method, which is achieved by minimizing the KL divergence of the classifier's prediction for various representations of the same original attribute $X$. Additional representations are obtained by adopting dropout when training the discriminator. This way, we are able to include an additional term in the loss function for training the auxiliary classifier, which is defined as follows:
\begin{equation}
    L_{d}(\psi,m,m')=D_{KL}(p_{\psi}(y|m),p_{\psi}(y|m') )
\label{eq:drop}    
\end{equation}

where $m$ is the extracted features without using dropout, and $m'$ denotes those generated by the discriminator with a low dropout rate. In this work, $y$ is a probabilistic output of the softmax distribution function. In addition, we make use of multiple cluster heads (softmax layers) that are described as follows:

\begin{equation}
\begin{split}
L (\psi;m) &= \frac{1}{C}\sum_{j=1}^C \frac{1}{N}\sum_{n=1}^N \frac{1}{2}{R_{SAT}(\psi;m_n,T(m_n))}
\\ &+ \frac{1}{2}L_{d}(\psi,m,m') + \lambda I_{\psi}(M,Y)
\end{split}
\label{eq:function_complete}
\end{equation}

where $C$ indicates the implemented number of cluster heads, $\lambda$ is a weight parameter, and $M$ is the new domain space produced by the discriminator. 
Instead of using a single cluster head, this work adopts multiple cluster heads together with the introduced tolerance factor, which is found to be able to significantly enhance the model's performance and robustness, in particular, when the tolerance factor $\delta$ is set to be small. This can be attributed to the fact that with the help of multiple cluster heads, the probabilistic outputs are simultaneously optimised with a variety of random parameters. In addition, we also included a few overcluster heads, whose outputs are larger than the predefined number $k$, in order to capture more details such as variations or sub-classes, within similar classes. Note that in this work, we assume $k$ is a hyper-parameter to be defined prior to the model training. \par 
By modifying the loss function and introducing multiple cluster heads, we are able to enhance the efficiency of the auxiliary classifier for clustering using the features generated by the discriminator, which is confirmed by our experimental results.
In the case of a high $k$ value, $\delta$-tolerance is initialised with a larger value, since we assume that training samples within sub-classes are not evenly distributed. The effects of the parameter setting of $\delta$ and the use of multiple cluster heads on the performance of the classifier are presented 

in Tables \ref{tab: effects cluster-heads} and \ref{tab:imbalanced}.\par

\SetKwInOut{Initialize}{Initialize}
\SetKwInOut{Generate}{Generate}

\begin{algorithm}[htb]
\KwData{$X=\{x_{i}\}_{i=1}^{n}$}
\Initialize{$(G_\theta,D_\theta,Aux_\theta)$ }
\While{Convergence condition not satisfied}{
  \hspace{0.7mm} \textbullet \hspace{0.1mm} Sample $n$ elements $\{x^{(1)},...,x^{(n)}\}$ \\
 \textbullet \hspace{0.1mm} Sample random minibatch $z_n \in [-1,1]$\\
\textbullet \hspace{0.1mm} Update the discriminator by ascending: \\
$\vcenter{
  \begin{align*}
  \nabla_{D_\theta} \frac{1}{n}\sum_{i=1}^{n}[log(D(x_i))+log(1-D(G(z_i)))]
 \end{align*}
 }$
  \textbullet \hspace{0.1mm} Sample random minibatch $z_m \in [-1,1]$\\
  \textbullet \hspace{0.1mm} Update the generator by ascending: \\
  $\vcenter{
  \begin{align*}
  \nabla_{G_\theta} \frac{1}{n}\sum_{i=1}^{n}[log(D(G(z_i)))]
 \end{align*}
 }$
}

\Generate{$D(X) \rightarrow M$}
\While{Convergence condition not satisfied}{
  \hspace{0.7mm} \textbullet \hspace{0.1mm} Sample $n$ features $\{m^{(1)},...,m^{(n)}\}$ \\
  \textbullet \hspace{0.1mm} Sample $n$ invariant features with a low dropout rate \\ \hspace{0.7mm} $\{{m'}^{(1)},...,{m'}^{(n)}\}$ \\
  \textbullet \hspace{0.1mm} Update the auxiliary classifier by ascending: \\
\begin{align*}
\begin{split}
\nabla_{Aux_\psi} &= \frac{1}{C}\sum_{j=1}^C \frac{1}{N}\sum_{n=1}^N\frac{1}{2}{R_{SAT}(\psi;m_n,T(m_n))}
\\ &+ \frac{1}{2}L_{d}(\psi,m,m') + \lambda I_{\psi}(M,Y)
\end{split}
\end{align*}

}
\caption{Proposed clustering process}
\label{alg:Process}
\end{algorithm}

Algorithm \ref{alg:Process} describes in detail the overall training process comprised of two consecutive learning phases. Note that neither stage in the proposed framework requires any labels for the data.

\section{Experiments}
\label{sec:results}

In this section we evaluate the proposed clustering method on a range of datasets and compare it to several state-of-the-art methods. We demonstrate that our method is capable of handling imbalanced datasets and classification tasks. Additionally we evaluate our method under different parameter settings and configurations, including the use of single or multiple cluster heads.

\subsection{Datasets}

We evaluate our algorithm on four popular benchmark image datasets, MNIST, CIFAR-10, CIFAR-100-20 and STL (a smaller subset of ImageNet). All relevant details in terms of the training samples, number of clusters and image resolution are presented in Table \ref{Dataset description}.

\begin{table}[htb]
\caption{A description of the datasets used in the experiments}
\resizebox{\columnwidth}{!}{
\begin{tabular}{lccc}
\hline
\textbf{Dataset} & \textbf{Training Samples} & \textbf{No. Clusters} & \textbf{Resolution} \\ \hline
\textbf{MNIST} & 70000 & 10 & 28x28x1 \\
\textbf{CIFAR-10} & 60000 & 10 & 32x32x3 \\
\textbf{CIFAR-100/20} & 60000 & 20 & 32x32x3 \\
\textbf{STL10*} & 113000 & 10 & 96x96x3 \\ \hline
\multicolumn{4}{l}
{
\begin{tabular}[c]{@{}p{8.6cm}@{}}*STL10 is evaluated only for the labelled subset of 13000 images.\end{tabular}}
\end{tabular}
}
\label{Dataset description}
\end{table}

In order to reduce the model's parameters, and consequently the duration of training phases, the following re-scales or crops have been applied. 1) MNIST training samples are center cropped to the dimensions of 24x24x1, 2) STL10 images are resized to 48x48x3, without additional transformation functions being applied. Furthermore, STL10 is evaluated only for the labelled subset whereas the model is trained across the complete dataset. For the CIFAR-100/20 dataset, we evaluated our proposed model based on $20$ super-classes also called 'coarse' across all experiments.

\subsection{Evaluation Metric}

In order to evaluate the performance of the proposed framework, we use a traditional metric function for clustering, namely absolute accuracy (ACC), which is expressed as follows:

\begin{dmath}
ACC = \max_{m \in M} \frac{ \sum_{i=1}^{N} \boldsymbol{1} [{l_i = m(c_i)}] }{N}
\end{dmath}

where $N$ denotes the number of testing samples, $l$ is the ground truth, $c$ is the model's assigned cluster, and $M$ is the set of all available one-to-one mappings. Upon the completion of the above metric, the accuracy is computed with the best matching between the cluster assignment and the given ground truth. This can be efficiently computed using the Hungarian algorithm \cite{doi:10.1002/nav.3800020109}.

\subsection{Implementation Details}

All experiments were based on an implementation of the DCGAN with a convolutional architecture. Here, we replace the ReLU \cite{Nair:2010:RLU:3104322.3104425} activation of the generator with the Leaky ReLU \cite{DBLP:journals/corr/XuWCL15}, where the parameter of the Leaky ReLU activation function is set to 0.2 for both nets. As designed in DCGAN, every convolutional operation is followed by a batch normalization layer in both nets of the framework. Additionally, similar to \cite{NIPS2014_5423}, we found it is beneficial to use dropout \cite{JMLR:v15:srivastava14a} to stabilize the training in the discriminator. The dropout rate is set to $20\%$ across all implementations. The last convolutional layer of the original discriminator in DCGAN is flattened and directly forwarded to the discriminator's probabilistic output. In our approach, an intermediate fully connected layer is employed between the last convolutional operation and the output layer of the discriminator to reduce the dimensionality of vector $M$. Further details regarding the discriminator' structure are presented in Table \ref{tab:implementation_details}, while the model's parameters for the corresponding structure are included in Table \ref{tab: model parameters}.  %\par
The generator model is developed as proposed in \cite{radford2015unsupervised} except for the last layer, where a convolutional operation transposes the input to the original height and width of the image prior to the last boundary output layer. 

The auxiliary classifier is based on a fully connected architecture. For the experimental results presented in Table \ref{tab:Unsupervised Clustering}, the auxiliary net consists of two hidden layers, and a ReLU activation is applied. The probabilistic outputs are determined via the softmax function. We implement five cluster heads in total in the experiments, and one over cluster head (i.e., with a larger $k$). Weight regularization is not implemented in this net. Inspired by \cite{DBLP:journals/corr/abs-1807-06653}, we generate a number of replications of the adversarial transformed features within each mini-batch in order to learn invariant representations of the data. 
We found that a higher number of replications leads to a faster convergence of the classifier model and increases the clustering performance.
In our experiments, the input data is replicated five times with alternative transformations and all six instances are propagated to the model in a single batch. The code is implemented in python using Tensorflow and can be made available upon request.

\begin{table}[tb]
\caption{The Convolutional Discriminator Architecture}
\begin{tabular}{ccc}
\hline
\textbf{\begin{tabular}[c]{@{}c@{}}MNIST\\-\\ 24x24x1\end{tabular}} & \textbf{\begin{tabular}[c]{@{}c@{}}CIFAR10-100/20\\-\\ 32x32x3\end{tabular}} & \textbf{\begin{tabular}[c]{@{}c@{}}STL\\-\\ 48x48x3\end{tabular}} \\ \hline 
\begin{tabular}[c]{@{}c@{}}1xConv\\ Kernel=4x4\\ Stride=2x2\\ Output=12x12x32\end{tabular} & \begin{tabular}[c]{@{}c@{}}1xConv\\ Kernel=4x4\\ Stride=2x2\\ Output=16x16x64\end{tabular} & \begin{tabular}[c]{@{}c@{}}1xConv\\ Kernel=4x4\\ Stride=2x2\\ Output=24x24x64\end{tabular} \vspace{2mm}\\ 
\begin{tabular}[c]{@{}c@{}}1xConv\\ Kernel=4x4\\ Stride=2x2\\ Output=6x6x64\end{tabular} & \begin{tabular}[c]{@{}c@{}}1xConv\\ Kernel=4x4\\ Stride=2x2\\ Output=8x8x128\end{tabular} & \begin{tabular}[c]{@{}c@{}}1xConv\\ Kernel=4x4\\ Stride=2x2\\ Output=12x12x128\end{tabular}\vspace{2mm} \\
\begin{tabular}[c]{@{}c@{}}1xConv\\ Kernel=4x4\\ Stride=2x2\\ Output=3x3x128\end{tabular} & \begin{tabular}[c]{@{}c@{}}1xConv\\ Kernel=4x4\\ Stride=2x2\\ Output=4x4x256\end{tabular} & \begin{tabular}[c]{@{}c@{}}1xConv\\ Kernel=4x4\\ Stride=2x2\\ Output=6x6x256\end{tabular} \vspace{2mm}\\
Flatten@1152 & FC@1024 &  FC@1024 \\ \hline
\multicolumn{3}{l}{
\begin{tabular}[c]{@{}p{8.6cm}@{}}*FC denotes a fully connected layer.\end{tabular}
}
\end{tabular}
\label{tab:implementation_details}
\end{table}

\begin{table}[h]
\caption{Parameter settings of the model}
\resizebox{\columnwidth}{!}{
\begin{tabular}{lcc}
\hline
\textbf{Dataset} & \textbf{Discriminator} & \textbf{Generator} \\ \hline
\textbf{MNIST} & 167K & 289K \\
\textbf{CIFAR-10/100} & 4.85M & 4.40M \\
\textbf{STL10*} & 10.10M & 11.99M \\ \hline
\multicolumn{3}{l}{
\begin{tabular}[c]{@{}p{8.6cm}@{}}'K' denotes thousands, and 'M' millions respectively.\end{tabular}
}
\end{tabular}
}
\label{tab: model parameters}
\end{table}

\subsection{Experimental Settings}

In this subsection, we discuss the settings for the hyper-parameters to be used in the experiments. We followed the recommendations in \cite{radford2015unsupervised} regarding weights initialization, the optimizer, and the mini-batch size in training the DCGAN. Specifically, the model parameters of the GAN are initialized with a Gaussian distribution of a standard deviation of $0.02$. Both the generator and discriminator are trained by Adam \cite{DBLP:journals/corr/KingmaB14} with a learning rate of $10^{-4}$ and $\beta_1=0.5$. Both the generator and discriminator are trained for a total of $10^3$ iterations. 
\par
For the clustering process, we set the weight parameter $\lambda$ in Equation  (\ref{eq:function_complete}) to $0.2$ across all experiments. The tolerance factor $\delta$ is defined to be $10^{-4} \cdot log(k)$, where $k$ is equal to the predefined number of clusters. The value of $\delta$ is chosen considering the imbalanced distribution within mini-batches. In the experiments, one overcluster head is used, for which we set $\delta=10^{-2} \cdot log(k')$, where $k'$ is set to be larger than the predefined number of clusters, since we expected that the sub-clusters are not evenly distributed. \par

For computing the adversarial transformation ($m+r_{adv}$), we set $\alpha_{r}$ of the first round of random perturbation to $0.3$ for CIFAR-10/100, and STL, and to $1.0$ for MNIST, respectively. In the second round of VAT process,  $\alpha_{adv}$ is set to $0.15$ for CIFAR-10/100 and STL, while it is set to $0.2$ for MNIST.

We set the dropout rate $m'$ in Equation (\ref{eq:drop}) to $10\%$. The auxiliary classifier is fine tuned with a mini-batch size of 500 training samples. The weight parameters of the auxiliary are initialized based on a Gaussian distribution with a standard deviation $10^{-3}$ for CIFAR-10/100, and $10^{-2}$ for STL and MNIST. Similar to the GAN, the Adam optimizer is selected for the training process with a learning rate of $10^{-4}$. The auxiliary classifier is trained for $10^{3}$ epochs in total in each experiment.

\begin{table*}[tb]
\caption{Comparative results on benchmark clustering problems.}
\begin{center}
%\resizebox{\columnwidth}{!}{
\begin{tabular}{lcccc}
%\multicolumn{5}{c}{\textbf{Table 4.3 Unsupervised Clustering Results}} \\ 
\hline
\textbf{Method} & \textbf{MNIST} & \textbf{CIFAR-10} & \textbf{CIFAR-100-20} & \multicolumn{1}{c}{\textbf{STL10}} \\ \hline
K-means & 53.49\% & 20.6\% & 12.97\% & 19.20\% \\
AE  \cite{NIPS2006_3048} \dag & 81.23\% & 31.35\% & 16.45\% & 30.3\% \\
VAR \cite{kingma2013autoencoding} \dag  & 83.17\% & 29.08\% & 15.17\% & 28.15\% \\
DEC \cite{pmlr-v48-xieb16}\ddag & 84.30\% & 30.10\% $\sharp$ & 18.50\% $\sharp$ & 35.90\% $\sharp$ \\
DCGAN \cite{radford2015unsupervised} \dag & 82.80\% & 31.50\% & 15.10\% & 29.80\% \\
{\bf DCGAN ours} \dag $\natural$ $\pm$ & 88.40\% & 52.36\% & 28.04\% & 43.91\% \\
JULE \cite{Yang2016JointUL} & 96.40\% & 27.15\% & 13.67\% & 27.69\% \\
DAC \cite{DBLP:conf/iccv/ChangWMXP17} & 97.75\% & \textbf{52.18\%} & \textbf{23.75}\% & 46.99\% \\
VADE \cite{DBLP:journals/corr/jiangZTTZ16}& 95.00\% & - & - & 84.45\% $\ast$  \\
IMSAT \cite{Hu2017} & 98.40\% & 45.60\% $\ast$ & 27.50\% $\ast$ & 94.10\% $\ast$\\
ADC \cite{DBLP:conf/dagm/HausserPGAC18} & \textbf{99.20\%} & 32.50\% & 18.90\% $\S$ & \textbf{53.00\%} \\
IIC \cite{DBLP:journals/corr/abs-1807-06653} & \textbf{99.20\%} & \textbf{61.70\%} & \textbf{25.70} & \textbf{59.60\%} \\ \hline
OURS (best) & \textbf{99.02\%} & \textbf{70.04\%} & \textbf{32.44\%} & \textbf{58.65\%} \\
OURS (avg) & 98.85\% $(\pm 0.14\%)$ & 69.22\% $(\pm 0.83\%)$ & 30.88\%  $(\pm 1.11\%)$ & 55.87\% $(\pm 1.81\%)$ \\ \hline
\multicolumn{5}{l}{
\begin{tabular}[l]{@{}p{12.6cm}@{}} \dag Methods perform feature extraction followed by K-means clustering. \\ 
\ddag Combination of a deep method and K-means. \\
$\sharp$ A pre-processing application of HOG is used. \\
$\natural$ Implementation of DCGAN with the application of Sobel filters prior to discrimination input. \\
$\ast$  Feature extraction is performed through a neural network model trained on Imagenet. \\
$\S$ Results obtained from table by  \cite{DBLP:journals/corr/abs-1807-06653}. \\
$\pm$ Proposed DCGAN architecture with a flattened hidden layer and Sobel filters.
\end{tabular}}
%\begin{tabular}[c]{@{}p{15.6cm}@{}}\dag Methods which firstly perform feature extraction and clustering follows via K-means. \ddag Combination of a deep method and K-means. $\sharp$ A pre-processing application of HOG is used. $\ast$  A feature extraction is performed through a neural network model trained on Imagenet. $\S$ Result is being obtained from table by  \cite{DBLP:journals/corr/abs-1807-06653}.\end{tabular}}
\end{tabular}
%}
\label{tab:Unsupervised Clustering}
\end{center}
\end{table*}

\subsection{Clustering Results}

In Table \ref{tab:Unsupervised Clustering}, we demonstrate the performance of the proposed method by comparing it with state-of-the-art methods. We report the accuracy scores of the best performed experiment of our method and the average performance with the corresponding standard deviation of five independent runs. Note that the reported predictions were made by the best performing cluster head having the lowest loss. All evaluations were made over the complete dataset except for STL10, which is evaluated only for the labelled subset. For convenience, the top three methods for each experiment are highlighted in bold font. From the results presented in Table \ref{tab:Unsupervised Clustering}, we can find that our approach exhibits competitive clustering performance compared to the state-of-the-art techniques. On some benchmark datasets, the proposed method outperforms all other methods under comparison. In particular, the proposed algorithm (average performance) reaches a margin of 7.5\% improvement on CIFAR-10 and 5\% on CIFAR-100 in comparison with the state-of-the-art. Recall that  the proposed method does not perform any direct image augmentation to the original data; instead, it relies on transformations that are not specific to image data. This leads to a notable advantage since the proposed strategy can be extended to non-image data. It should be pointed out that although VADE and IMSAT outperform all other algorithms under comparison on the STL10 dataset, they both use features extracted by a model that is pre-trained using supervised learning on Imagenet. 
\par

\begin{figure*}[t]
    \centering
    \includegraphics[width=1 \textwidth]{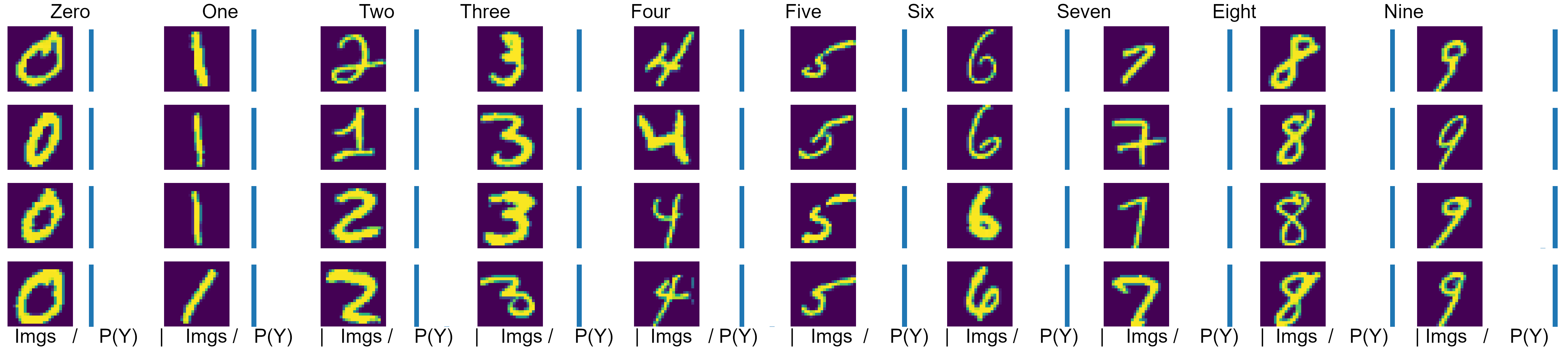}
    \caption{Clustering results on CIFAR-10 and MNIST. Visualization of the predictions made by the proposed clustering method, which includes the relevant images and their class probabilities (P(Y)).}
    \label{fig:multiple images MNIST}
\end{figure*}

Figure \ref{fig:multiple images MNIST} illustrates the results of the proposed algorithm (indicated by the relevant probabilities predicted by the model) for CIFAR-10 and MNIST experiments. As can be observed, on the MNIST dataset (lower panel), our model successfully assigned the right label on the majority of the corresponding images, although one inaccurate prediction is made for class `two' (third column), where an element from class `one' with some common spatial lines were picked. By looking at the results on the CIFAR-10 dataset (upper panel), we find that our model is able to distinguish the classes surprisingly well with few cases of inaccurate predictions. We see that clusters with common patterns, i.e., 'Automobile' with 'Truck' and 'Horse' with 'Deer' are successfully clustered. The lowest performance can be seen on class 'Cat' which represents an uncertainty of the model confidence. We present the individual clustering results on the CIFAR-10 dataset in Table \ref{tab:imbalanced}. 
Figure \ref{fig:TSNE emb} shows the 2D scatter plot of the representations $M$ for the CIFAR-10 dataset, where the left panel is based  the ground truth and the right panel is the clustering results of our model. From these results, we can conclude that the clusters identified by the auxiliary classifier are very close to the ground truth.

\begin{figure*}[t]
    \centering
    \includegraphics[width=1 \textwidth]{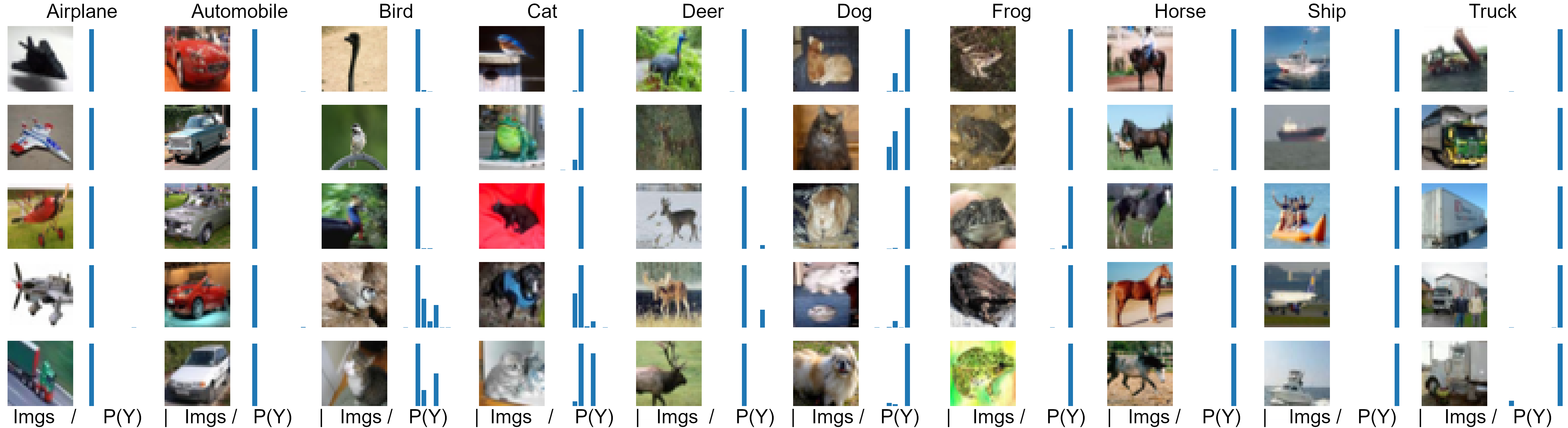}
%    \caption{Clustering results on CIFAR-10 dataset. Visualization of random predictions is made by our proposed clustering strategy which includes the relevant image and probabilities.}
    \label{fig:multiple images CIFAR10}
\end{figure*}

\begin{figure*}[t]
    \centering
    \includegraphics[width=1 \textwidth]{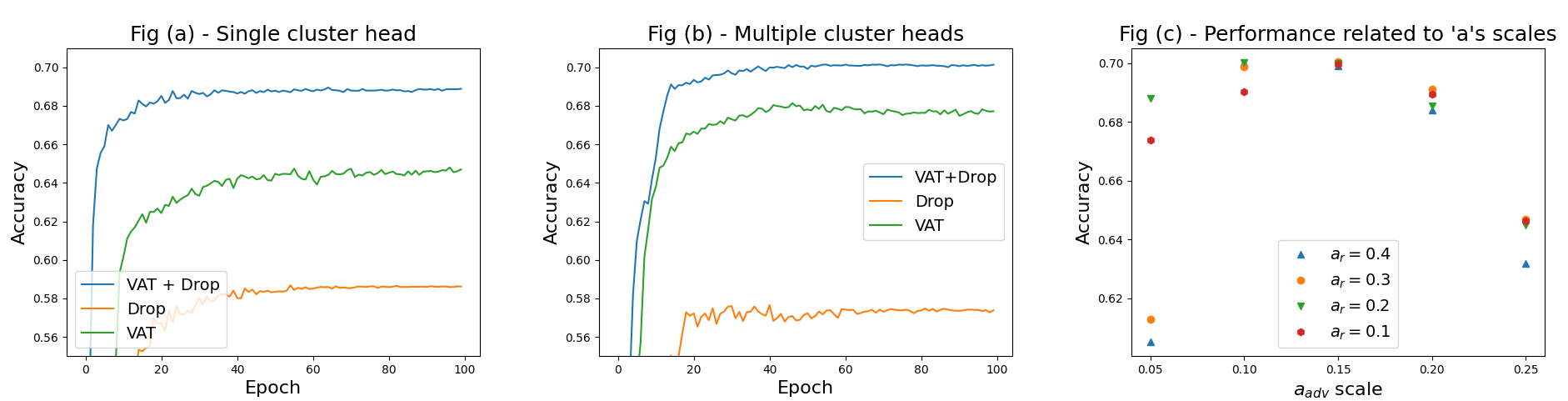}
    \caption{(a) and (b) present the accuracy of the three variants of the loss function based on a single cluster head model and multi-cluster heads, respectively. (c) illustrates the change of performance over different values of two scale factors $a_r$ and $a_{adv}$. For convenience $a_r$ is coloured and marked with different shapes, where the $x$ axis indicates the taken values of $a_{adv}$.}
    \label{fig:plots}
\end{figure*}

\subsection{Ablation Studies on Auxiliary Classifier}
\label{sub:auxiliary classifier}

To demonstrate the advantage of the proposed auxiliary classifier, and to evaluate the behaviour of each regularization component, we conduct a range of experiments based on Equation \ref{eq:function_complete} on the CIFAR-10 dataset. In the first group of evaluations, the following settings apply across all experiments: the deployed model uses a single cluster head and the MI (Equation \ref{eq:minimize mi}) part of Equation \ref{eq:function_complete} remains unchanged. We compare the following three variants of the regularization term: 1) the VAT component (Equation \ref{eq:rsat}) is switched on and the dropout part (Equation \ref{eq:drop}) is switched off; 2) the VAT component is switched off and the dropout part is switched on; and 3) Both components are switched on using the default parameters in Equation \ref{eq:function_complete}. In the second group of experiments, we use a set of five cluster heads for each model and one over-cluster head. The training methodology remains the same as in the first group. In all experiments, the scale factor remains constant as $\delta=10^{-4} \cdot \text{log}(k)$ for all single-cluster heads and $\delta=10^{-2} \cdot \text{log}(k')$ for the over-cluster head if this is involved. Note that for the VAT evaluation of the single-cluster head we decrease the value of $a_{adv}$ from $0.15$ to $0.1$. The results obtained for the single-cluster head and the multiple cluster heads are presented in Fig. \ref{fig:plots} (a) and Fig. \ref{fig:plots} (b), respectively. As can be observed from these plots, the use of both VAT and dropout terms performs the best.

\begin{table*}[tb!]

%\resizebox{\columnwidth}{!}{
\centering
\caption{Predicted accuracy per class in CIFAR-10}
\begin{tabular}{l|ccccc|cccc}
\multicolumn{1}{c}{} & \multicolumn{5}{c}{\textbf{constant $\delta$-tolerance}} &  \multicolumn{4}{c}{\textbf{variable $\delta$-tolerance}} \\ \hline
\textbf{Dropped} & \textbf{0\%} & \textbf{10\%} & \textbf{20\%} & \textbf{30\%} & \textbf{40\%} & \textbf{10\%} & \textbf{20\%} & \textbf{30\%} & \textbf{40\%}\\ \hline
\textbf{airplane} &78.40\% & 75.31\% & 78.15\% & 77.90\% & 77.31\%&

76.22\% & 76.94\% & 73.55\% & 74.47\% \\
\textbf{automobile} & 89.38\% & 88.93\% & 88.73\% & 88.64\% & 89.25\% &

88.67\% & 89.58\% & 90.12\% & 89.03\% \\
\textbf{bird} & 50.72\% & 52.87\% & 41.83\% & 39.38\% & 11.56\% &

45.43\% & 44.96\% & 9.40\% & 10.61\% \\
\hline
\textbf{cat} & 23.43\% & 43.73\% & 20.35\% & 23.40\% & 40.38\% &

24.30\% & 25.68\% & 44.97\% & 44.85\% \\
\textbf{deer} & 72.03\% & 71.02\% & 70.88\% & 71.87\% & 72.55\% &

70.55\% & 71.50\% & 72.52\% & 72.97\% \\
\textbf{dog} & 59.65\% & 50.65\% & 51.73\% & 47.68\% & 45.52\% &

59.08\% & 56.62\% & 49.42\% & 47.48\% \\
\textbf{frog} & 84.50\% & 83.20\% & 84.40\% & 83.77\% & 83.63\% &

84.17\% & 83.72\% & 83.75\% & 82.95\% \\
\textbf{horse} & 69.93\% & 69.88\% & 69.73\% & 69.77\% & 69.03\% &

70.52\% & 69.62\% & 67.12\% & 67.47\% \\
\textbf{ship} & 89.72\% & 89.00\% & 89.35\% & 90.50\% & 90.20\% &

89.88\% & 89.15\% & 88.90\% & 88.20\% \\
\textbf{truck} & 82.72\% & 81.88\% & 82.67\% & 81.78\% & 81.33\% & 

83.13\% & 82.98\% & 82.90\% & 82.92\% \\ \hline
\multicolumn{10}{l}{\begin{tabular}[c]{@{}p{13.2cm}@{}}
Clustering performance of the auxiliary classifier with dropped features from the following classes: Airplane, Automobile and Bird. The dropout rates are mentioned on top of each column (where $10\%$ represents 600 elements).
\end{tabular}}
\label{tab:imbalanced}
\end{tabular}
%}
\end{table*}

Additionally, we examine the impact of the scale factors $a_r$ and $a_{adv}$ on the final performance of the model. They both are hyper-parameters indicating the impact of the produced adversarial noise on the representations. The performance is illustrated in Fig. \ref{fig:plots} (c) for $a_{r} \in \{0.1 , 0.2, 0.3 , 0.4\}$ and $a_{adv} \in \{0.05 , 0.10, 0.15 , 0.20 ,0.25\}$. In these figures, the range of $a_{r}$ is denoted by different shapes and colours and $a_{adv}$ is the $x$-axis. From these results, we see the performance of the model is relatively insensitive to the perturbation noise when $a_{adv}$ is in the range of $[0.1, 0.2]$ and it achieves the best robustness when $r_{adv} = 0.15 $.

As the last part of experiments on the hyper-parameters, we compare the auxiliary classifier with three of its variants on the CIFAR-10 dataset and the results are presented in Table \ref{tab: effects cluster-heads}. The first variant has one single cluster head only and is trained by minimizing the proposed objective function with $\delta=0$. The second variant also has a single cluster head, however, $\delta=10^{-4} \cdot  \text{log}(k)$. The third and fourth variants of the proposed auxiliary classifier have five cluster heads with $k=10$ outputs and one additional over-cluster heads with $k'=50$ outputs. The main difference between the third and fourth is that the former is trained with $\delta=0$ for all cluster heads, while for the latter, $\delta=10^{-4} \cdot \text{log}(k)$ for the five cluster heads with ten outputs and $\delta=10^{-2} \cdot \text{log}(k')$ for the one overcluster heads with $50$ outputs. Note that the fourth variant is the standard one used in all the previous experiments and its parameter setting are listed in Table \ref{tab:Unsupervised Clustering}. The softmax activation is applied for all probabilistic outputs in the set of experiments here. All experiments have taken place in the same generated domain. The auxiliary classifier has two hidden layers, each consisting of 1024 units. The mini-batch size is set to 500 in each training round. In all the above four variants, the classifier is optimized for 200 iterations on the full dataset. Three independent runs are performed and the best performances are reported in Table \ref{tab: effects cluster-heads}. Note that for the variants with multiple cluster heads, we present the accuracy of the best cluster head, together with the mean accuracy averaged over all heads and the lowest performance.

\begin{table}[htb]
\caption{Effects of multiple cluster-heads and $\delta$-tolerance}
\centering
%\resizebox{\columnwidth}{!}{
\begin{tabular}{lcccc}
%\multicolumn{5}{c}{\textbf{Table 4.4 Auxuliary Classifier Results on CIFAR-10}} \\ 
\hline
\multicolumn{1}{c}{\multirow{2}{*}{\textbf{Method}}} & \multirow{2}{*}{\textbf{Single Output}} & \multicolumn{3}{c}{\textbf{Multiple Outputs}} \\ \cline{3-5} 
\multicolumn{1}{c}{} &  & \textbf{Best} & \multicolumn{1}{c}{\textbf{Average}} & \multicolumn{1}{c}{\textbf{Lowest}} \\ \hline
\textbf{$\delta=0$} & 65.71\% & 66.59\% & 63.81\% & 57.47\% \\
\textbf{$\delta>0$} & 68.20\% & 70.04\% & 69.64\% & 68.68\% \\ \hline
%\multicolumn{5}{l}{\begin{tabular}[c]{@{}p{7.6cm}@{}}
%Comparison of clustering capability between the original %IMSAT and our proposed adjustment.
%\end{tabular}}
\end{tabular}
\label{tab: effects cluster-heads}
%}
\end{table}

From the results in Table \ref{tab: effects cluster-heads}, we can conclude that the auxiliary classifier proposed in this work with multiple cluster heads and $\delta>0$ in the loss function has resulted in the best performance among all four compared variants. Meanwhile, the classifiers with multiple heads performed better than those with a single head. Note, however, that the variants with a single cluster head demonstrate lower performance on $\delta>0$, in comparison with the multiple heads variants. These results confirm the benefit of using multiple clusters and modifying the loss function.

\subsection{Evaluation on Imbalanced Data}

We further evaluated the effectiveness of the proposed auxiliary classifier on imbalanced datasets and the results are listed in Table \ref{tab:imbalanced}. Note that CIFAR-10 is a perfectly balanced dataset, where each class contains exactly 6000 training samples. Specifically, we examine the influence of the tolerance factors on the performance of the auxiliary classifier by randomly dropping some data from the three classes: 'airplane', 'automobile' and 'bird'. The dropping rate is increased by $10\%$ in each experiment as follows: $0\%$, i.e., no data is dropped; $10\%$, with 600 data pairs being dropped; $20\%$, 1200; $30\%$, 1800; and $40\%$, 2400. Two sets of experiments are conducted, one with $\delta$ being fixed, and  the other with $\delta$ being varied. Detailed settings of the hyper-parameters are described below. 

\begin{itemize}
    \item \textbf{Constant tolerance rates}. In this set of experiments, the tolerance rate is kept constant while the dropping rate is changed. Note that for the cluster heads with 10 outputs, %$\delta=0.0001$ and for the overcluster heads with 50 outputs, $\delta=0.01$.
    $\delta=10^{-4}$ and for the overcluster heads with 50 outputs, $\delta=10^{-2}$. 
    \item \textbf{Variable tolerance rates}. In this set of experiments, the tolerance factors vary together with the drop rate. The drop rate, the tolerance for the cluster heads, and the tolerance rate for the overcluster heads are: 
    \textbf{$10\%$}, $10^{-4}$, $2\cdot10^{-2}$; \textbf{$20\%$}, $10^{-3}$, $2\cdot10^{-2}$; \textbf{$30\%$}, $10^{-3}$, $5\cdot10^{-2}$; and \textbf{$40\%$}, $2\cdot10^{-3}$, $5\cdot10^{-2}$, respectively.
    %\textbf{$10\%$}, $0.0001$, $0.02$; \textbf{$20\%$}, $0.001$, $0.02$; \textbf{$30\%$}, $0.001$, $0.05$; and \textbf{$40\%$}, $0.002$, $0.05$, respectively.
\end{itemize}

The classifier consists of five cluster-heads with 10 outputs and two over cluster-heads with $50$ outputs. In Table \ref{tab:imbalanced}, the results of the best-performing cluster-head after 200 iterations are presented. \par

As we can see from Table \ref{tab:imbalanced}, the auxiliary classifier is considerably robust to the imbalanced data. The accuracy slightly deteriorates in the cases with data being dropped. More specifically, the worst performance is observed for the 'bird' class, as the accuracy achieved on this class was the lowest among the three dropped classes, even without removing any data. Furthermore, as presented in Fig. \ref{fig:TSNE emb} (A), classes 'cat' and 'bird' share some similarities in the generated features, and therefore the performance of these two classes is similar when the drop rate increases. It is surprising to see that the accuracy is improved when the drop rate is set to $10\% \text{ or } 40\%$, whereas the performance remains unchanged for the settings. From these results, we can conclude that the overall performance of the classifier is satisfactorily high on all imbalanced datasets. Additionally, consistently high accuracy is observed across all classes, although, surprisingly, an extra term in the loss function for the tolerance has made little impact on the accuracy. 

\begin{figure*}[t]
    \centering
    \includegraphics[width=1\textwidth]{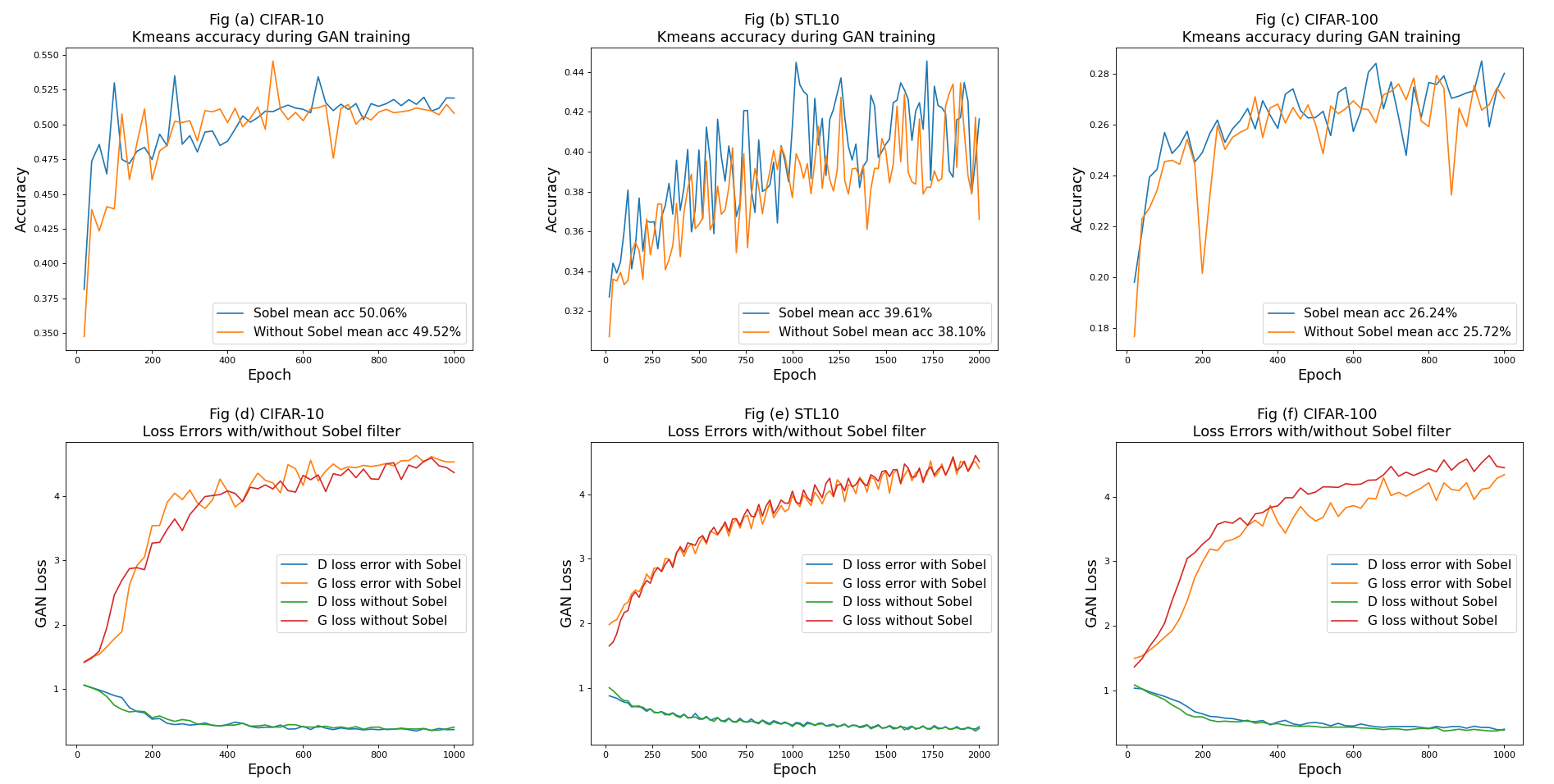}
    \caption{Comparison of the discriminator's performance with or without the Sobel operation. The plots in the top row show the absolute accuracy achieved with the features extracted by k-means calculated in every 20 epochs (the blue line indicates the Sobel filter operation). The bottom row visualizes the errors of the discriminator and the generator, implying that adding the Sobel filters in the discriminator does not impair the training performance.}
    \label{fig:comparison sobel}
\end{figure*}
\subsection{Feature Extraction Validation}
\label{sub:Feature Extraction Validation}

\subsubsection{Influence of Sobel filter} 

Figure \ref{fig:comparison sobel} illustrates the benefit of integrating a Sobel filter in the discriminator for feature extraction. Here, we implement two similar architectures, one with and the other without a Sobel filter. Firstly, we can observe from the plots in the bottom row that the additional constant filters have no effect on the already competitive training phase of the GAN framework across the three training sets. The plots in the top row visualize the accuracy achieved using the features extracted by k-means in every 20 epochs. The dimension of the extracted representations is reduced to 20 with PCA. The higher margin between the architectures can be attributed to the challenging nature of the dataset (STL10) with a mean accuracy of $39.61\%$ and $38.10\%$, respectively, with and without the Sobel operator. Note that the algorithm without the Sobel filters requires $33.4 \text{sec}$ for each epoch on STL10 on Nvidia RTX 2080Ti, and an additional $0.4 \text{sec}$ is needed when the Sobel filters is used.

\subsubsection{Classification} 
In the previous results, we have show the promising performance of the proposed model for data clustering. Here, we want to demonstrate the capability of the modified discriminator in the proposed model for classification. To do so, we evaluated our approach by comparing it with a list of benchmark representation learning techniques, namely autoencoders, the original DCGAN with an intermediate hidden flatten layer similar to our proposed architecture, a bidirectional GAN model, and the Deep InfoMax. Note that none of the compared methods relies on image augmentation operations. Instead, the representation is learnt directly from the original training set. 

All experiments have been conducted on CIFAR-10, which consists of 50000 training and 10000 testing samples. Similar to \cite{hjelm2019learning}, the produced features were evaluated against the following two approaches: 

\begin{itemize}
  \item Linear-SVM. Firstly, the generated features of the flatten layer are reduced to $50$ components via PCA. Then, a linear L2-SVM is trained on this reduced domain.
  \item Non-Linear model. A single hidden layer network of 200 units is built on top of the main framework. During the training, the weight parameters are optimized separately from the main framework. A dropout technique is introduced to alleviate over-fitting in the training phase with a dropout rate of $40\%$.
\end{itemize}

\begin{figure*}[t]
    \centering
    \includegraphics[width=1\textwidth]{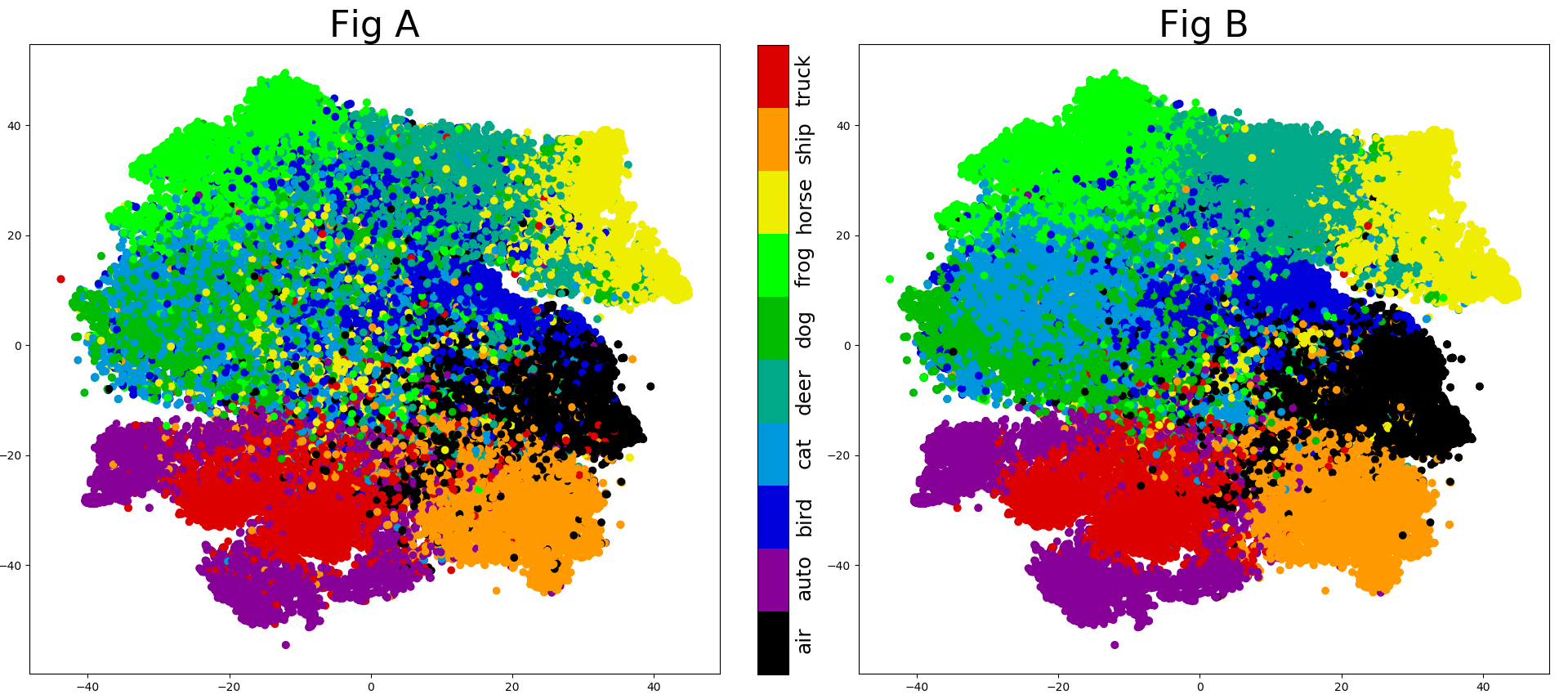}
    \caption{2D visualizations of the CIFAR-10 dataset via t-SNE. The left panel is colored based on the ground truth and the right panel is based on the predictions of the auxiliary classifier. For convenience, 'airplane' is notated as 'air', and 'automobile' as 'auto'.}
    \label{fig:TSNE emb}
\end{figure*}

\begin{table}[htb]
\caption{Classification of Extracted Features}
%\resizebox{\linewidth}{!}{
\centering
\begin{tabular}{lx{2cm}x{3cm}}
\hline 
 & \multicolumn{2}{c}{\textbf{CIFAR-10}}  \\ \hline
\textbf{Model} & \textbf{L2-SVM} & \textbf{Non-Linear Model} \\ \hline
\textbf{VAE} & 42.58\% & 58.61\%   \\
\textbf{AE} & 40.58\% & 55.02\%    \\
\textbf{AAE \dag} & 43.34\% & 57.19\%  \\
\textbf{BiGAN \dag} & 38.42\% & 62.74\%   \\
\textbf{DIM(L) \dag} & 54.06\% & 75.57\%  \\
\textbf{DCGAN \ddag} & 54.70\% & 70.61\%  \\
\textbf{DCGAN $\ast$} & 61.06\% & 74.45\%  \\
\textbf{DCGAN $\ast$ $\sharp$} & \textbf{71.22\%} & \textbf{78.25\%} \\ \hline
\multicolumn{3}{l}{ \begin{tabular}[c]{@{}p{8.0cm}@{}}  \ddag denotes the initial DCGAN architecture evaluated only in the last convolutional layer. \\ $\ast$ DCGAN architectures with a fully connected hidden layer prior to the probabilistic output. \\ $\sharp$ indicates that the Sobel filter is applied. \\ \dag Due to the high processing demand of the above experiments, the results of BiGAN, DIM(L) and AAE are taken from \cite{hjelm2019learning}.\end{tabular}}
\end{tabular}
\label{tab:supervised classification}
%}
\end{table}

Table \ref{tab:supervised classification} presents the prediction results of the two techniques on the testing data. From these results, we can see that our approach exhibits a superior performance on both classification tasks, where a significant margin is observed %\textbf{What does 'via' mean? - changed it to 'via use of'}
when using the linear SVM model. 
%Notably, the result of the auxiliary classifier in Table \ref{tab:Unsupervised Clustering} achieved its best accuracy of $70.04\%$ in the unsupervised manner, outperforming the majority of the supervised method under comparison. % \textbf{Double check if the above changes make sense}
Notably, the result of the auxiliary classifier in Table \ref{tab:Unsupervised Clustering} achieved its best accuracy of $70.04\%$ in unsupervised manner, outperforming the majority of the methods under comparison in Table \ref{tab:supervised classification}. In these compared methods, feature extraction is accomplished via unsupervised learning followed by the classification task, reached in a supervised manner with the relevant labels to be provided during the training. Finally, we can observe from these results that the Sobel filter is able to enhance the classification accuracy of the linear SVM by a large margin of $10\%$.  

\subsubsection{Visualization} 
In this phase, we initially reduce the generated features ($M$) via PCA, %\textbf{I do no understand what 'likewise' means here - changed to 'equivalently to'} 
equivalently to the classification task, and select only the $50$ components with the largest variance. The data reduced by PCA is further reduced into a two-dimensional embedding space via t-SNE, a state-of-the-art visualization technique \cite{vanDerMaaten2008}.

The results of t-SNE are presented in Fig. \ref{fig:TSNE emb} (A). The visualization is colored according to the ground truth label, where each color represents one particular class. In addition, Table \ref{tab:imbalanced} presents the prediction accuracy on CIFAR-10 when the tolerance rate varies, and when a certain percentage of the data is dropped. By observing the results in Table \ref{tab:imbalanced}  and the visualization in Fig. \ref{fig:TSNE emb} (A) (which is based on ground truth), 

We can note that there is a correlation between the visualized data and classifier's predictions. From Fig. \ref{fig:TSNE emb} (A), we recognize that similar classes are placed in the neighbourhood in the scatter plot. For example, 'automobile' (notated as 'auto', purple) and 'truck' (red), 'deer' (light green) and 'horse' (yellow). Additionally, the subsets of 'ship' and 'airplane' share the same cyan background, indicating that 'airplane' instances are placed in the 'ship' cluster neighbourhood. On the other hand, classes such as 'dog' (dark green), 'cat' (cyan), and 'bird' (dark blue) are mixed in the scatter plot. Therefore, the performance of the auxiliary classifier heavily relies on the quality of the features generated by the discriminator. Refer also to the top panel of Fig. \ref{fig:multiple images MNIST}  for the predicted results on CIFAR-10.

\section{Conclusion}

In this work, we present a deep framework for image clustering. Experimental results show very competitive performance of the proposed framework compared with state-of-the-art deep clustering techniques. Even with a single cluster head, the proposed framework is able to achieve surprisingly high accuracy, and with multiple cluster heads, the performance is further enhanced. Finally, we demonstrate the robustness of the proposed framework for clustering imbalanced data.

Despite the promising results, the effectiveness of the auxiliary classifier in the proposed framework highly relies on the separability of the features generated by the discriminator. Additionally, large-scale GAN models may suffer from unstable performance during the training process, as well as the high computational cost for the optimization of the generator. Thus, our future work will investigate new techniques for reducing the computational cost. In addition, we intend to extended the proposed framework to other types of high-dimensional data such as audio and spectral data, since its internal mechanisms for data augmentation do not rely on any image augmentation techniques.

%\ifCLASSOPTIONcompsoc
%\section*{Acknowledgments}
%\else
%\section*{Acknowledgment}
%\fi

%\ifCLASSOPTIONcaptionsoff
%  \newpage
%\fi

\bibliographystyle{IEEEtran}
\bibliography{ref}

\end{document}